\documentclass{article}
\usepackage{spconf,amsmath,graphicx,hyperref}
\usepackage{amssymb}
\usepackage{bm}
\usepackage{xurl}
\usepackage{pifont}
\usepackage{siunitx}
\usepackage{multirow}
\usepackage{booktabs}

\newcommand{\myparagraph}[1]{%
  \par\vspace{0.7ex}%
  \noindent\textbf{#1}\hspace{0.5em}%
}

\ninept

\title{TinyCardioUNet: IMU-to-ECG Translation with Graph-Encoded Inter-Axis Dependencies and Tensor Decomposition-Based Parameter Reduction}
\name{Seungwoo Han, Ingon Chanpornpakdi, Motoi Noda, Puwadej Leelasiri, Ibuki Hiruma, Toshihisa Tanaka* \thanks{*Corresponding author: Toshihisa Tanaka, tanakat@cc.tuat.ac.jp \\ This study was partly supported by the JST CRONOS project JPMJCS24K7.}}
\address{Department of Electrical Engineering and Computer Science,\\
Tokyo University of Agriculture and Technology, Japan \\
\{han, mint, noda23, paul, hiruma25\}@sip.tuat.ac.jp, tanakat@cc.tuat.ac.jp}
\begin{document}
%
\maketitle
\begin{abstract}
Estimating electrocardiography (ECG) from a chest-worn inertial measurement unit (IMU) enables continuous heart rate (HR) monitoring without the discomfort of electrodes.
We propose TinyCardioUNet, a lightweight UNet that uses all six IMU axes without prior channel selection, refines its bottleneck with a graph neural network that encodes inter-axis dependencies, and employs tensor decomposition with automatic variational Bayesian rank selection for parameter reduction.
On a public dataset, TinyCardioUNet achieves an RMSE of $0.098$ and a Pearson correlation coefficient of $0.677$ with only $36.0$k parameters and remains comparatively robust to additive noise, demonstrating accurate ECG reconstruction with a compact model.

\end{abstract}
\section{Introduction}
\label{sec:intro}

Cardiovascular disease (CVD) remains the leading cause of death worldwide, accounting for roughly 19.8 million deaths per year \cite{who2025cvd}, and chronic stress is a major modifiable contributor to its onset. 
Key physiological markers of both are heart rate (HR) and heart rate variability (HRV).
Because an elevated resting HR and impaired HRV regulation are associated with high stress and an increased risk of CVD \cite{dekker2000}, continuous HR monitoring is necessary for stress management and the early prevention of heart disease. 

Electrocardiography (ECG), which records the electrical activity of the heart, is the clinical gold standard for HR monitoring \cite{Fye1994}, but its reliance on gel electrodes causes physical discomfort and skin irritation during long-term wear \cite{martinez-tabares2014, taji2014}. As promising solutions for electrode-free HR monitoring, seismocardiography (SCG) \cite{bozhenko1961} and gyrocardiography (GCG) \cite{jafaritadi2017}, which record cardiac vibrations and rotational dynamics using chest-worn inertial measurement units (IMU), have attracted attention. Beyond HR monitoring \cite{lahdenoja2016heart}, SCG and GCG have also been applied to the diagnosis of conditions such as heart failure \cite{mehrang2020}, and chest-worn IMUs have emerged as a supporting modality for both wearable devices and continuous clinical monitoring.

To combine the convenience of IMU with the interpretability of ECG, recent studies have used deep learning to translate chest-worn IMU signals into ECG \cite{tapotee2024, Skoric2025}. Skoric \textit{et al.} \cite{Skoric2025} proposed an ECG reconstruction model architecture using conditional generative adversarial networks (CGANs) with a six-axis IMU. Its generator alone contains approximately 13.6M parameters. Such a large chest-worn IMU-to-ECG translation model requires substantial computational resources, making it difficult to deploy on resource-constrained edge hardware (e.g., Raspberry Pi Pico series).
Tapotee \textit{et al.} \cite{tapotee2024} proposed the deeply-supervised spatial-attention UNet to translate SCG and GCG waveforms into ECG, but their best-performing model used only two of the six available IMU axes ($\mathrm{SCG}_z, \mathrm{GCG}_y$), as the remaining axes degraded ECG estimation due to motion artifacts. However, relying on a fixed subset of empirically selected axes may limit robustness when axis-specific signal quality varies across subjects or measurement conditions.
Thus, novel methodologies are needed to leverage all IMU axes without prior channel selection while reducing computational costs without compromising reconstruction fidelity under noisy conditions. 

To address these issues, we propose TinyCardioUNet, a novel chest-worn IMU-to-ECG translation model that integrates inter-axis dependencies among IMU axes into the UNet \cite{ronneberger2015} architecture. The underlying idea behind our architecture is two-fold.
First, building on our earlier accelerometer-only model \cite{han2025}, which showed that modeling inter-sensor relationships captures global context that per-channel convolutions miss,
we encode inter-axis dependencies among all six SCG/GCG axes as a graph and refine the UNet bottleneck with a graph neural network. Unlike conventional convolution-based approaches that treat channels independently or share filters across channels, this graph-based formulation explicitly models the structured inter-axis dependencies inherent in IMU measurements, enabling information from a noise-corrupted axis to be complemented by its physically coupled axes.
Second, we compress the model parameters via tensor decomposition, yielding a lightweight architecture suited for deployment on resource-constrained edge devices.

\section{Materials and Methods}
\begin{figure*}[t!]
    \centering
    \includegraphics[width=\textwidth]{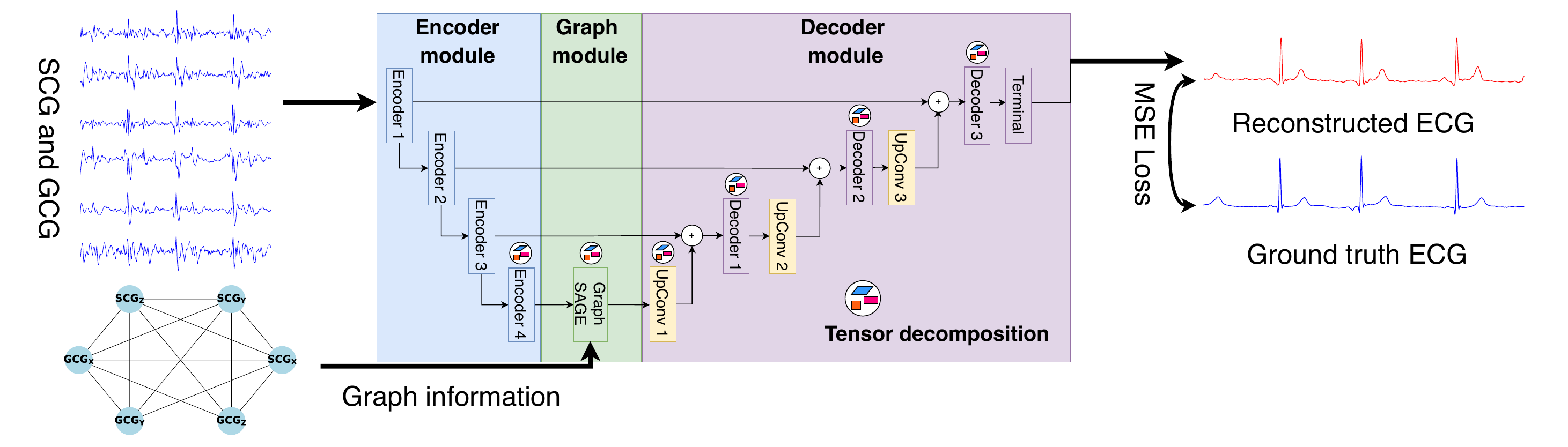}
    \caption{Overview of the TinyCardioUNet pipeline. The pipeline is divided into three stages: model training with ground-truth (GT) signals and graph using mean squared error (MSE) loss; model compression via tensor decomposition; and inference to reconstruct the ECG waveform.}
    \label{fig:cardioUnet_framework}
\end{figure*}

\subsection{Dataset}
We use the ``Mechanocardiograms with ECG Reference'' dataset \cite{kaisti2019}, which contains simultaneous SCG, GCG, and ECG recordings from 29 healthy male volunteers, acquired at 800 Hz for up to 10 minutes per volunteer. Across these volunteers, age was $29 \pm 5$ years, height $179 \pm 5$ cm, weight $76 \pm 11$ kg, and BMI $24 \pm 3$ kg/m\textsuperscript{2}. Altogether, the recordings amounted to roughly 260 minutes. SCG and GCG were obtained from an IMU secured to the sternum, with the lateral, head-to-foot, and dorso-ventral directions defined as the x, y, and z-axes. ECG was recorded in a Lead II configuration. Measurements were taken in the supine and the left and right lateral decubitus positions.

\subsection{Data preprocessing}
We follow the preprocessing pipeline of Tapotee \textit{et al.}\ \cite{tapotee2024}, since we use the same dataset and apply the same experimental protocol. Volunteers numbered 11, 21, and 24 form the test set and the remaining 26 volunteers form the training set. The raw data is downsampled to \SI{256}{Hz}, and a sixth-order Butterworth bandpass filter is applied (cutoffs of \SI{1}{Hz}--\SI{40}{Hz} for SCG and GCG, \SI{0.1}{Hz}--\SI{40}{Hz} for ECG), followed by a seventh-order polynomial baseline correction to remove residual baseline wandering. The training data are segmented into windows of 1024 samples with $50\%$ overlap, and each window is z-score standardized and then min-max normalized to $[0,1]$.

\subsection{TinyCardioUNet architecture}

As illustrated in Fig.~\ref{fig:cardioUnet_framework}, TinyCardioUNet follows the encoder--bottleneck--decoder structure of UNet \cite{ronneberger2015} and consists of three parts: an encoder module, a graph module, and a decoder module.

\myparagraph{Encoder module}
Each encoder block applies a 1D convolution (Conv1D), batch normalization (BN), and a rectified linear unit (ReLU). A max-pooling operation (Pool, kernel $2$, stride $2$) is inserted between successive blocks, halving the temporal length at each step so that an input of length $N$ is reduced to $N/2$, $N/4$, and finally $N/8$. The four encoder blocks use Conv1D layers with output channels of 16, 32, 64, and 128, respectively.

\myparagraph{Graph module}
The bottleneck feature map is refined by a single GraphSAGE layer
\cite{hamilton2017} with a mean aggregator. Each of the $C$ bottleneck
channels is treated as a node $v$, whose length-$(N/8)$ temporal vector
$h_v$ is the node feature. The first six nodes correspond to the six IMU
axes $(\mathrm{SCG}_x,\mathrm{SCG}_y,\mathrm{SCG}_z,
\mathrm{GCG}_x,\mathrm{GCG}_y,\mathrm{GCG}_z)$ and form a complete
graph; the remaining nodes have no incident edges. A complete graph
models the coupled cardiac motion across all six axes without assuming
a priori sparsity. Using a fixed, row-normalized adjacency matrix
$\hat{\mathbf{A}}=(\hat{A}_{vu})\in\mathbb{R}^{C\times C}$, neighborhood aggregation is
\begin{equation}
h_{\mathcal{N}(v)}
= \sum_{u\in\mathcal{V}}\hat{A}_{vu}h_u
= \frac{1}{|\mathcal{N}(v)|}
  \sum_{u\in\mathcal{N}(v)}h_u .
\end{equation}
The neighborhood and self features are transformed separately as:
\begin{equation}
h_v'=\mathbf{W}_1h_v+\mathbf{W}_2h_{\mathcal{N}(v)}+\mathbf{b},
\end{equation}
followed by temporal $\ell_2$ normalization:
\begin{equation}
    h_v'\leftarrow \frac{h_v'}{\|h_v'\|_2}.
\end{equation}
No nonlinearity is applied, keeping $\mathbf{W}_1$ and $\mathbf{W}_2$
as linear layers that can be factorized in the compression stage.

\myparagraph{Decoder module}
The decoder restores the bottleneck features to the input resolution through three upsampling stages with output widths 64,
32, and 16, respectively. Each stage performs nearest-neighbor upsampling by a factor of two, followed by a Conv1D--BN--ReLU block. After each stage, the feature map is concatenated with the corresponding encoder feature map through a skip connection and passed through a further convolutional block. A terminal $1 \times 1$ Conv1D projects the $16$-channel feature map to a single channel, whose linear output is the reconstructed ECG. All of TinyCardioUNet's convolutions use kernel size $3$, stride $1$, and padding $1$, except the terminal $1 \times 1$ convolution (stride $1$, padding $0$).

\myparagraph{Training}
The model is trained for $60$ epochs with a batch size of $256$ using the evolved sign momentum optimizer \cite{chen2023} at a learning rate of $1 \times 10^{-3}$. We minimize the mean squared error (MSE) between the reconstructed and GT ECG segments.

\subsection{Model compression using tensor decomposition}
\label{subsec:td}
Tensor decomposition (TD) factorizes a multi-way data array into a set of
lower-dimensional factors and is widely used for neural network compression
\cite{Kolda2009, kim2015, cao2017tensor, Dai2025}. We compress the convolutional and graph layers of
the proposed model with Tucker-based decomposition, automatically selecting
each layer's rank via variational Bayesian matrix factorization (VBMF)
\cite{Nakajima2013}. Since our model is built on
1D convolutions, each
kernel is a third-order tensor $\mathcal{K} \in \mathbb{R}^{D \times S \times T}$,
where $D$ is the kernel size and $S$ and $T$ denote the numbers of input and
output channels, respectively. Tucker-2 decomposition factorizes the kernel
along the input and output channel modes only, leaving the kernel mode intact:
\begin{equation}
  \mathcal{K}_{d,s,t}
  = \sum_{r_2=1}^{R_2} \sum_{r_3=1}^{R_3}
    \mathcal{O}_{d,r_2,r_3}\
    U^{(2)}_{s,r_2}\
    U^{(3)}_{t,r_3}
\end{equation}
where $d$ is the kernel-size mode index, $\mathcal{O} \in \mathbb{R}^{D \times R_2 \times R_3}$ is the core tensor, $U^{(2)} \in \mathbb{R}^{S \times R_2}$ and $U^{(3)} \in \mathbb{R}^{T \times R_3}$ are the factor matrices, and $R_2 \le S$, $R_3 \le T$ are the Tucker ranks. Substituting this decomposition replaces a single convolution with three
consecutive ones. We apply Tucker-2 decomposition to an empirically selected subset of the encoder and decoder blocks. For the graph module, the two GraphSAGE transformation weights $\mathbf{W}_1$
and $\mathbf{W}_2$ are matrices, so we apply Tucker-1 decomposition, which is
equivalent to the truncated singular value decomposition (SVD). Each
$W \in \mathbb{R}^{S \times T}$ is factorized into two low-rank factors,
replacing the layer with two smaller linear maps. Following \cite{kim2015}, each layer's rank is obtained automatically from the
global analytic solution of empirical VBMF applied to the mode-wise unfoldings
of its weight tensor, capped at $90\%$ of the original dimension to guarantee a
non-trivial reduction. Table \ref{tab:td} shows the network parameters of TinyCardioUNet and whether tensor decomposition was applied. We adopt a post-training compression scheme: the model
is first trained to
convergence, its trained weights are decomposed with the selected ranks, and the
compressed model is fine-tuned (FT) for $5$ or $10$ epochs at a learning rate of
$2 \times 10^{-4}$ to recover the reconstruction accuracy lost during decomposition.

\begin{table}[t!]
\centering
\caption{TinyCardioUNet parameters and structure.}
\label{tab:td}
\resizebox{0.5\textwidth}{!}{%
\begin{tabular}{lcccc}
\hline
\textbf{Block name} & \textbf{Size $(D,S,T)$} & \textbf{Structure} & \textbf{Output shape} & \textbf{TD} \\\hline
Encoder1 & $(3,6,16)$ & Conv1D--BN--ReLU & ($B$, 16, $N$) & -- \\ 
Encoder2 & $(3,16,32)$ & Pool--Conv1D--BN--ReLU & ($B$, 32, $N/2$) & -- \\ 
Encoder3 & $(3,32,64)$ & Pool--Conv1D--BN--ReLU & ($B$, 64, $N/4$) & -- \\ 
Encoder4 & $(3,64,128)$ & Pool--Conv1D--BN--ReLU & ($B$, 128, $N/8$) & \ding{51} \\ 
GraphSAGE & $(-,128,128)$ & GraphSAGE & ($B$, 128, $N/8$) & \ding{51} \\ 
UpConv1 & $(3,128,64)$ & Upsample--Conv1D--BN--ReLU & ($B$, 64, $N/4$) & \ding{51} \\ 
Decoder1 & $(3,128,64)$ & Conv1D--BN--ReLU & ($B$, 64, $N/4$) & \ding{51} \\ 
UpConv2 & $(3,64,32)$ & Upsample--Conv1D--BN--ReLU & ($B$, 32, $N/2$) & -- \\ 
Decoder2 & $(3,64,32)$ & Conv1D--BN--ReLU & ($B$, 32, $N/2$) & \ding{51} \\ 
UpConv3 & $(3,32,16)$ & Upsample--Conv1D--BN--ReLU & ($B$, 16, $N$) & -- \\ 
Decoder3 & $(3,32,16)$ & Conv1D--BN--ReLU & ($B$, 16, $N$) & \ding{51} \\ 
Terminal layer & $(1,16,1)$ & Conv1D & ($B$, 1, $N$) & -- \\\hline
\multicolumn{5}{l}{$D$: kernel size, $S$: input channel size, $T$: output channel size} \\
\multicolumn{5}{l}{$B$: batch size, $N$: input signal length}
\end{tabular}
}
\end{table}

\section{Results}
\begin{table*}[t!]
\centering
\caption{Performance comparison across noise conditions (No noise, SNR $\{20, 15, 10\}$\ dB). Values are mean$\pm$standard deviation. All results under our experimental setup.}
\label{tab:benchmark}
\resizebox{\textwidth}{!}{%
\begin{tabular}{l ccc ccc ccc ccc}
\hline
\multirow{2.5}{*}{\textbf{Model / Metric}}
 & \multicolumn{3}{c}{\textbf{No noise}} & \multicolumn{3}{c}{\textbf{SNR 20}} & \multicolumn{3}{c}{\textbf{SNR 15}} & \multicolumn{3}{c}{\textbf{SNR 10}} \\
\cmidrule(lr){2-4}\cmidrule(lr){5-7}\cmidrule(lr){8-10}\cmidrule(lr){11-13}
 & MAE$\downarrow$ & RMSE$\downarrow$ & PCC$\uparrow$
 & MAE$\downarrow$ & RMSE$\downarrow$ & PCC$\uparrow$
 & MAE$\downarrow$ & RMSE$\downarrow$ & PCC$\uparrow$
 & MAE$\downarrow$ & RMSE$\downarrow$ & PCC$\uparrow$ \\\hline
CGAN \cite{Skoric2025}
 & 0.156\tiny{$\pm$0.047} & 0.200\tiny{$\pm$0.050} & 0.190\tiny{$\pm$0.059}
 & 0.162\tiny{$\pm$0.046} & 0.207\tiny{$\pm$0.049} & 0.179\tiny{$\pm$0.059}
 & 0.173\tiny{$\pm$0.043} & 0.222\tiny{$\pm$0.047} & 0.158\tiny{$\pm$0.057}
 & 0.207\tiny{$\pm$0.038} & 0.264\tiny{$\pm$0.043} & 0.117\tiny{$\pm$0.052} \\
UNet \cite{ronneberger2015}
 & 0.205\tiny{$\pm$0.092} & 0.230\tiny{$\pm$0.082} & 0.555\tiny{$\pm$0.186}
 & 0.203\tiny{$\pm$0.091} & 0.227\tiny{$\pm$0.081} & 0.551\tiny{$\pm$0.181}
 & 0.199\tiny{$\pm$0.086} & 0.226\tiny{$\pm$0.076} & 0.532\tiny{$\pm$0.170}
 & 0.210\tiny{$\pm$0.069} & 0.239\tiny{$\pm$0.060} & 0.455\tiny{$\pm$0.151} \\
WaveNet \cite{oord2016wavenet}
 & 0.118\tiny{$\pm$0.065} & 0.143\tiny{$\pm$0.067} & 0.597\tiny{$\pm$0.169}
 & 0.123\tiny{$\pm$0.063} & 0.150\tiny{$\pm$0.063} & 0.567\tiny{$\pm$0.161}
 & 0.132\tiny{$\pm$0.057} & 0.159\tiny{$\pm$0.058} & 0.521\tiny{$\pm$0.151}
 & 0.160\tiny{$\pm$0.063} & 0.187\tiny{$\pm$0.061} & 0.388\tiny{$\pm$0.119} \\
WaveUNet \cite{stoller2018wave}
 & 0.203\tiny{$\pm$0.101} & 0.219\tiny{$\pm$0.096} & 0.528\tiny{$\pm$0.222}
 & 0.213\tiny{$\pm$0.107} & 0.229\tiny{$\pm$0.102} & 0.507\tiny{$\pm$0.227}
 & 0.238\tiny{$\pm$0.120} & 0.254\tiny{$\pm$0.115} & 0.459\tiny{$\pm$0.230}
 & 0.296\tiny{$\pm$0.108} & 0.310\tiny{$\pm$0.101} & 0.343\tiny{$\pm$0.196} \\\hline
TinyCardioUNet (baseline, ours)
 & 0.093\tiny{$\pm$0.055} & 0.123\tiny{$\pm$0.056} & 0.622\tiny{$\pm$0.208}
 & 0.101\tiny{$\pm$0.058} & 0.128\tiny{$\pm$0.058} & 0.616\tiny{$\pm$0.208}
 & 0.118\tiny{$\pm$0.064} & 0.144\tiny{$\pm$0.063} & 0.584\tiny{$\pm$0.215}
 & 0.194\tiny{$\pm$0.074} & 0.215\tiny{$\pm$0.069} & 0.453\tiny{$\pm$0.197} \\
TinyCardioUNet w/o graph module (ours)
 & 0.210\tiny{$\pm$0.056} & 0.246\tiny{$\pm$0.050} & 0.380\tiny{$\pm$0.106}
 & 0.202\tiny{$\pm$0.060} & 0.242\tiny{$\pm$0.056} & 0.378\tiny{$\pm$0.117}
 & 0.174\tiny{$\pm$0.052} & 0.217\tiny{$\pm$0.051} & 0.396\tiny{$\pm$0.130}
 & 0.158\tiny{$\pm$0.036} & 0.201\tiny{$\pm$0.037} & 0.355\tiny{$\pm$0.133} \\
TinyCardioUNet + TD (ours)
 & 0.091\tiny{$\pm$0.042} & 0.119\tiny{$\pm$0.042} & 0.527\tiny{$\pm$0.170}
 & 0.092\tiny{$\pm$0.041} & 0.120\tiny{$\pm$0.041} & 0.517\tiny{$\pm$0.168}
 & 0.098\tiny{$\pm$0.043} & 0.126\tiny{$\pm$0.043} & 0.483\tiny{$\pm$0.172}
 & 0.105\tiny{$\pm$0.046} & 0.136\tiny{$\pm$0.045} & 0.409\tiny{$\pm$0.171} \\
TinyCardioUNet + TD + FT@5epoch (ours)
 & 0.071\tiny{$\pm$0.043} & 0.102\tiny{$\pm$0.046} & \textbf{0.679}\tiny{$\pm$0.179}
 & 0.073\tiny{$\pm$0.044} & 0.102\tiny{$\pm$0.046} & \textbf{0.684}\tiny{$\pm$0.172}
 & 0.076\tiny{$\pm$0.045} & 0.103\tiny{$\pm$0.046} & \textbf{0.684}\tiny{$\pm$0.169}
 & \textbf{0.099}\tiny{$\pm$0.047} & \textbf{0.123}\tiny{$\pm$0.048} & \textbf{0.634}\tiny{$\pm$0.159} \\
TinyCardioUNet + TD + FT@10epoch (ours)
 & \textbf{0.068}\tiny{$\pm$0.046} & \textbf{0.098}\tiny{$\pm$0.047} & 0.677\tiny{$\pm$0.177}
 & \textbf{0.070}\tiny{$\pm$0.047} & \textbf{0.099}\tiny{$\pm$0.048} & 0.680\tiny{$\pm$0.169}
 & \textbf{0.074}\tiny{$\pm$0.048} & \textbf{0.102}\tiny{$\pm$0.049} & 0.672\tiny{$\pm$0.171}
 & 0.102\tiny{$\pm$0.061} & 0.127\tiny{$\pm$0.059} & 0.606\tiny{$\pm$0.162} \\\hline
\end{tabular}}
\end{table*}

\begin{table}[t!]
\centering
\caption{Computational cost comparison.}
\label{tab:cost}
\resizebox{\columnwidth}{!}{%
\begin{tabular}{l cc}
\hline
\textbf{Model} & \#Param.$\downarrow$ & FLOPs$\downarrow$ \\\hline
CGAN \cite{Skoric2025}          & 13.6M$^\text{A}$ & 614.11M$^\text{A}$ \\
UNet \cite{ronneberger2015}     & 10.8M            & 2{,}545.68M       \\
WaveNet \cite{oord2016wavenet}  & 48{,}257         & 98.17M            \\
WaveUNet \cite{stoller2018wave} & 3.4M             & 293.81M           \\\hline
TinyCardioUNet (baseline, ours)          & 131{,}361       & 68.89M        \\
TinyCardioUNet w/o graph module (ours)   & 98{,}465        & 56.28M        \\
TinyCardioUNet + TD + FT (ours)               & \textbf{36{,}004} & \textbf{27.99M} \\\hline
\multicolumn{3}{l}{$^\text{A}$ Generator parameters only.} \\
\end{tabular}}
\end{table}

\subsection{Evaluation metrics}

We assess the models in terms of waveform fidelity, computational cost, and heart rate agreement.

\myparagraph{Waveform fidelity}
Between the reconstructed signal $x$ and the GT ECG $y$ we report the mean absolute error (MAE), root-mean-squared error (RMSE), and Pearson correlation coefficient (PCC). MAE and RMSE are defined as follows:
\begin{equation}
\text{MAE} = \frac{1}{N}\sum_{i=1}^{N}|x_i - y_i|,\quad  \text{RMSE} = \sqrt{\frac{1}{N}\sum_{i=1}^{N}(x_i - y_i)^2}
\end{equation}
where $N$ is the number of time points in a segment and $i$ indexes a time point. All metrics are computed for each test segment and reported as the mean$\pm$standard deviation across all test segments. PCC is defined as follows:
\begin{equation}
 \text{PCC} = \frac{\sum_{i}(x_i - \bar{x})(y_i - \bar{y})}{\sqrt{\sum_{i}(x_i - \bar{x})^2}\,\sqrt{\sum_{i}(y_i - \bar{y})^2}}
\end{equation}
where $x_i$ and $y_i$ are the $i$-th time points of the reconstructed and GT ECG segments, and $\bar{x}$, $\bar{y}$ are their segment means. PCC is computed for each segment at zero lag. A cross-correlation lag search confirmed that this zero-lag choice fell within $\pm2$ samples ($\pm7.8$\,ms, $<1\%$ of a typical RR interval) of the population-optimal lag, so our estimates remained effectively in phase with the GT ECG. We also inject independent additive white Gaussian noise into each of the six IMU input channels of the test data to probe the robustness of TinyCardioUNet. For each segment, the noise variance is calibrated from that segment's own signal power so that the signal-to-noise ratio $\text{SNR}_{\text{dB}} = 10\log_{10}(P_{\text{signal}}/P_{\text{noise}})$ holds exactly at $\{20, 15, 10\}$\,dB.

\myparagraph{Computational cost} We further quantify model complexity by the parameter count ($\#$Param.) and the number of floating-point operations per segment (FLOPs) during model inference using FLOPpy \cite{Scala2026}.

\noindent
\textbf{Heart rate agreement} We compute the PCC between the average HR over four seconds of the ECG generated from our proposed model and the actual average HR from the GT ECG, defined as:
\begin{equation}
 \text{PCC}_{\text{HR}} = \frac{\sum_{i}(HR_i - \overline{HR})(\widehat{HR}_i - \overline{\widehat{HR}})}{\sqrt{\sum_{i}(HR_i - \overline{HR})^2}\,\sqrt{\sum_{i}(\widehat{HR}_i - \overline{\widehat{HR}})^2}}
\end{equation}
where $HR_i$ and $\widehat{HR}_i$ denote the ground-truth and estimated average HR over the $i$-th four-second window, respectively.

\subsection{Ablations and benchmark comparison}
\myparagraph{Effect of the graph module} 
Table~\ref{tab:benchmark} shows that inserting the bottleneck GraphSAGE layer raises PCC across all noise levels and lowers MAE and RMSE in all but the heaviest-noise (SNR 10 dB) condition. Because the two models share an identical encoder–decoder, this improvement is attributable to the graph module and temporal $\ell_2$ normalization, which enable explicit inter-axis information exchange beyond conventional convolutional processing.

\myparagraph{Effect of tensor decomposition} VBMF-ranked Tucker-2 and SVD decomposition compress the full model by 72.6\% in parameters and 59\% in FLOPs (Table~\ref{tab:cost}). Fine-tuning the decomposed model recovers and surpasses the uncompressed accuracy across all noise levels, with a small trade-off between the two schedules: the schedule with 10 epochs gives the lowest waveform error in clean and moderate-noise conditions, whereas the one with five epochs gives the highest PCC at every noise level and degrades least under heavy noise. We attribute this improvement to a regularization effect, as the low-rank structure constrains model capacity and mitigates overfitting \cite{cao2017tensor}.

\begin{figure*}[t!]
    \centering
    \includegraphics[width=\linewidth]{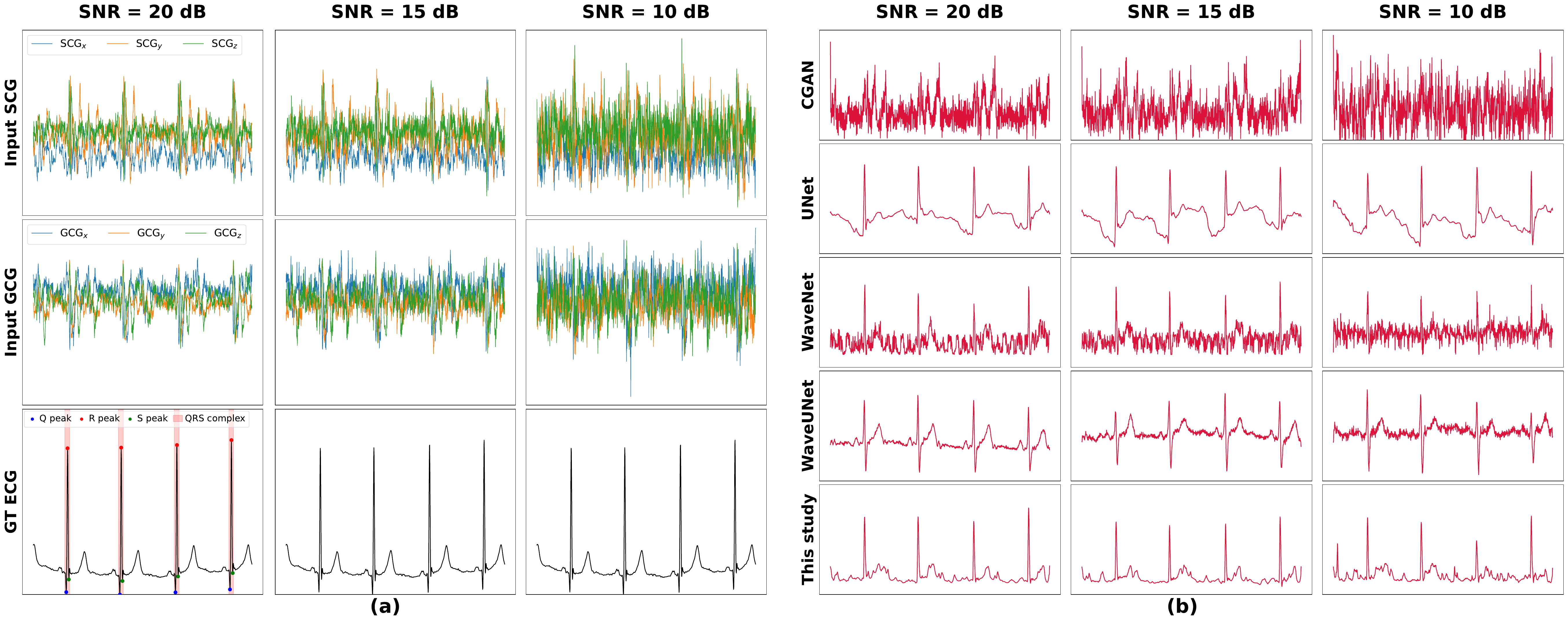}
    \caption{Examples of the ECG reconstruction from various models. (a) Input three-axis SCG and three-axis GCG at SNR $\{20, 15, 10\}$\ dB, with the GT ECG shown for reference. (b) Output ECG signals reconstructed by CGAN, UNet, WaveNet, WaveUNet, and TinyCardioUNet (TD + FT@10epoch) under the same SNR conditions.}
    \label{fig:result_vis}
\end{figure*}

\myparagraph{Comparison with various benchmarks}
We selected CGAN~\cite{Skoric2025}, UNet~\cite{ronneberger2015}, WaveNet~\cite{oord2016wavenet}, and WaveUNet~\cite{stoller2018wave} as benchmark models, all trained with a learning rate of $1 \times 10^{-3}$. For WaveUNet, however, this learning rate caused the training loss to diverge, so we adopted $1 \times 10^{-4}$ as proposed in the original paper. All benchmark models were optimized using the same optimizer, for the same number of epochs, and with the same batch size as TinyCardioUNet.
Compared with WaveNet, TinyCardioUNet (TD + FT@10epoch) lowers no-noise RMSE by 31\% and raises PCC from 0.597 to 0.677 while using fewer parameters and roughly $3.5\times$ fewer FLOPs. Against CGAN, UNet, and WaveUNet, it reduces no-noise RMSE by 51--57\% and substantially increases PCC, while using roughly two orders of magnitude fewer parameters. Figure~\ref{fig:result_vis} presents the reconstructed ECG for each model under varying IMU noise levels SNR $\{20, 15, 10\}$\ dB, shown for a representative segment (volunteer 11). CGAN degrades to nearly the noise level, suggesting that GAN-based generation requires carefully tuned hyperparameters to faithfully reconstruct the ECG. UNet, WaveNet, and WaveUNet capture the QRS complexes but suffer from baseline drift or invalid waveform generation, whereas TinyCardioUNet recovers the QRS complexes and their timing across all three SNR levels.
\section{Discussion}
TinyCardioUNet uses all six IMU axes and remains robust across noise levels with only 36.0k parameters. From no noise to SNR 10 dB, its PCC (TD + FT@10epoch) declines only from 0.677 to 0.606, whereas the far larger CGAN, UNet, WaveNet, and WaveUNet models show substantially greater degradation. This suggests that explicit inter-axis coupling and low-rank structure contribute to both parameter reduction and reconstruction accuracy.

Tapotee \textit{et al.}~\cite{tapotee2024} reported an MAE of $0.054\pm0.026$, an RMSE of $0.080\pm0.034$, and a $\text{PCC}_{\text{HR}}$ of $0.941$ under no noise, compared with our results of $0.068\pm0.046$, $0.098\pm0.047$, and $0.814$, respectively.
Their model, however, used the empirically selected best-performing SCG and GCG axes, with the channel combination selected based on test-set performance, whereas TinyCardioUNet uses all six IMU axes without channel selection.
Moreover, their input channels were selected based on noise characteristics without evaluation under varying-noise conditions; thus, their robustness under controlled input-noise levels was not assessed.

Nevertheless, this study has limitations that warrant further work. The dataset lacks diversity, as the experiment used healthy males rather than cardiac patients, and the proposed graph structure may not be optimal, leaving room for more effective domain-specific knowledge.
\section{Conclusion}
TinyCardioUNet is a lightweight UNet that uses all six IMU axes without prior channel selection and models their inter-axis dependencies via a bottleneck GraphSAGE layer. Compressed using VBMF-ranked tensor decomposition and brief fine-tuning, TinyCardioUNet achieves substantial reductions in parameters and computational cost while retaining competitive reconstruction accuracy.
These results highlight the potential of graph-based inter-axis modeling for resource-constrained multichannel physiological signal translation.

\vfill\pagebreak
\bibliographystyle{IEEEbib}
\bibliography{references}

@article{bozhenko1961,
  title = {{Seismocardiography--a new method in the study of functional conditions of the heart}},
  author = {Bozhenko, B. S.},
  year = 1961,
  month = sep,
  journal = {Terapevticheskii Arkhiv},
  volume = {33},
  pages = {55--64},
  issn = {0040-3660},
  langid = {russian},
  pmid = {13872234}
}

@inproceedings{chen2023,
  title = {Symbolic {{Discovery}} of {{Optimization Algorithms}}},
  author = {Chen, Xiangning and Liang, Chen and Huang, Da and Real, Esteban and Wang, Kaiyuan and Pham, Hieu and Dong, Xuanyi and Luong, Thang and Hsieh, Cho-Jui and Lu, Yifeng and Le, Quoc V.},
  year = 2023,
  month = dec,
  booktitle = {Advances in Neural Information Processing Systems},
  volume = {36},
  pages = {49205--49233},
  langid = {english},
}

@inproceedings{hamilton2017,
  title = {Inductive {{Representation Learning}} on {{Large Graphs}}},
  booktitle = {Advances in {{Neural Information Processing Systems}}},
  author = {Hamilton, Will and Ying, Zhitao and Leskovec, Jure},
  year = 2017,
  volume = {30},
}

@inproceedings{han2025,
  title = {A Lightweight Multi-Feature Fusion Deep Learning Architecture for Human {{ECG}} Reconstruction from Chest-Worn Accelerometer},
  booktitle = {2025 {{International Technical Conference}} on {{Circuits}}/{{Systems}}, {{Computers}}, and {{Communications}} ({{ITC-CSCC}})},
  author = {Han, Seungwoo and Chanpornpakdi, Ingon and Leelasiri, Puwadej and Noda, Motoi and Makanae, Sota and Irimajiri, Mami and Tanaka, Toshihisa},
  year = 2025,
  month = jul,
  pages = {1--5},
  address = {Seoul, Korea, Republic of},
  doi = {10.1109/ITC-CSCC66376.2025.11137726},
  isbn = {979-8-3315-5363-0},
  langid = {english},
}

@article{jafaritadi2017,
  title = {Gyrocardiography: {{A New Non-invasive Monitoring Method}} for the {{Assessment}} of {{Cardiac Mechanics}} and the {{Estimation}} of {{Hemodynamic Variables}}},
  shorttitle = {Gyrocardiography},
  author = {Jafari Tadi, Mojtaba and Lehtonen, Eero and Saraste, Antti and Tuominen, Jarno and Koskinen, Juho and Ter{\"a}s, Mika and Airaksinen, Juhani and P{\"a}nk{\"a}{\"a}l{\"a}, Mikko and Koivisto, Tero},
  year = 2017,
  month = jul,
  journal = {Scientific Reports},
  volume = {7},
  number = {1},
  pages = {6823},
  issn = {2045-2322},
  doi = {10.1038/s41598-017-07248-y},
  langid = {english},
}

@article{kaisti2019,
  title = {Stand-{{Alone Heartbeat Detection}} in {{Multidimensional Mechanocardiograms}}},
  author = {Kaisti, Matti and Tadi, Mojtaba Jafari and Lahdenoja, Olli and Hurnanen, Tero and Saraste, Antti and P{\"a}nk{\"a}{\"a}l{\"a}, Mikko and Koivisto, Tero},
  year = 2019,
  month = jan,
  journal = {IEEE Sensors Journal},
  volume = {19},
  number = {1},
  pages = {234--242},
  issn = {1558-1748},
  doi = {10.1109/JSEN.2018.2874706},
}

@inproceedings{martinez-tabares2014,
  title = {Very Long-Term {{ECG}} Monitoring Patch with Improved Functionality and Wearability},
  booktitle = {2014 36th {{Annual International Conference}} of the {{IEEE Engineering}} in {{Medicine}} and {{Biology Society}}},
  author = {{Martinez-Tabares}, F.J. and {Gaviria-Gomez}, N. and {Castellanos-Dominguez}, G.},
  year = 2014,
  month = aug,
  pages = {5964--5967},
  issn = {1558-4615},
  doi = {10.1109/EMBC.2014.6944987},
}

@inproceedings{ronneberger2015,
  title = {U-{{Net}}: {{Convolutional Networks}} for {{Biomedical Image Segmentation}}},
  shorttitle = {U-{{Net}}},
  booktitle = {Medical {{Image Computing}} and {{Computer-Assisted Intervention}} -- {{MICCAI}} 2015},
  author = {Ronneberger, Olaf and Fischer, Philipp and Brox, Thomas},
  editor = {Navab, Nassir and Hornegger, Joachim and Wells, William M. and Frangi, Alejandro F.},
  year = 2015,
  pages = {234--241},
  publisher = {Springer International Publishing},
  address = {Cham},
  doi = {10.1007/978-3-319-24574-4_28},
  isbn = {978-3-319-24574-4},
  langid = {english},
}

@article{tapotee2024,
  title = {{{M2ECG}}: {{Wearable Mechanocardiograms}} to {{Electrocardiogram Estimation Using Deep Learning}}},
  shorttitle = {{{M2ECG}}},
  author = {Tapotee, Malisha Islam and Saha, Purnata and Mahmud, Sakib and Alqahtani, Abdulrahman and Chowdhury, Muhammad E. H.},
  year = 2024,
  journal = {IEEE Access},
  volume = {12},
  pages = {12963--12975},
  issn = {2169-3536},
  doi = {10.1109/ACCESS.2024.3353463},
  langid = {american},
}

@article{mehrang2020,
  title = {{Classification of Atrial Fibrillation and Acute Decompensated Heart Failure Using Smartphone Mechanocardiography: A Multilabel Learning Approach}},
  volume = {20},
  ISSN = {2379-9153},
  DOI = {10.1109/jsen.2020.2981334},
  number = {14},
  journal = {IEEE Sensors Journal},
  publisher = {Institute of Electrical and Electronics Engineers (IEEE)},
  author = {Mehrang,  Saeed and Lahdenoja,  Olli and Kaisti,  Matti and Tadi,  Mojtaba Jafari and Hurnanen,  Tero and Airola,  Antti and Knuutila,  Timo and Jaakkola,  Jussi and Jaakkola,  Samuli and Vasankari,  Tuija and Kiviniemi,  Tuomas and Airaksinen,  Juhani and Koivisto,  Tero and Pankaala,  Mikko},
  year = {2020},
  month = jul,
  pages = {7957–7968}
}

@article{dekker2000,
  title = {{Low Heart Rate Variability in a 2-Minute Rhythm Strip Predicts Risk of Coronary Heart Disease and Mortality From Several Causes: The ARIC Study}},
  volume = {102},
  ISSN = {1524-4539},
  DOI = {10.1161/01.cir.102.11.1239},
  number = {11},
  journal = {Circulation},
  publisher = {Ovid Technologies (Wolters Kluwer Health)},
  author = {Dekker,  Jacqueline M. and Crow,  Richard S. and Folsom,  Aaron R. and Hannan,  Peter J. and Liao,  Duanping and Swenne,  Cees A. and Schouten,  Evert G.},
  year = {2000},
  month = sep,
  pages = {1239–1244}
}

@article{taji2014,
  title = {{Impact of Skin–Electrode Interface on Electrocardiogram Measurements Using Conductive Textile Electrodes}},
  volume = {63},
  ISSN = {1557-9662},
  DOI = {10.1109/tim.2013.2289072},
  number = {6},
  journal = {IEEE Transactions on Instrumentation and Measurement},
  publisher = {Institute of Electrical and Electronics Engineers (IEEE)},
  author = {Taji,  Bahareh and Shirmohammadi,  Shervin and Groza,  Voicu and Batkin,  Izmail},
  year = {2014},
  month = jun,
  pages = {1412–1422}
}

@inproceedings{kim2015,
  title={{Compression of Deep Convolutional Neural Networks for Fast and Low Power Mobile Applications}},
  author={Kim, Yong-Deok and Park, Eunhyeok and Yoo, Sungjoo and Choi, Taelim and Yang, Lu and Shin, Dongjun},
  booktitle={4th International Conference on Learning Representations (ICLR)},
  year = {2016}
}

@article{Kolda2009,
  title = {{Tensor Decompositions and Applications}},
  volume = {51},
  ISSN = {1095-7200},
  url = {http://dx.doi.org/10.1137/07070111X},
  DOI = {10.1137/07070111x},
  number = {3},
  journal = {SIAM Review},
  publisher = {Society for Industrial & Applied Mathematics (SIAM)},
  author = {Kolda,  Tamara G. and Bader,  Brett W.},
  year = {2009},
  month = Aug,
  pages = {455–500}
}

@article{Nakajima2013,
  title={{Global analytic solution of fully-observed variational Bayesian matrix factorization}},
  author={Nakajima, Shinichi and Sugiyama, Masashi and Babacan, S Derin and Tomioka, Ryota},
  journal={The Journal of Machine Learning Research},
  volume={14},
  number={1},
  pages={1--37},
  year={2013},
  publisher={JMLR. org}
}

@article{Scala2026,
  title = {{FLOPpy: A hardware-agnostic Python library to monitor the computational cost of machine and deep learning algorithms}},
  volume = {35},
  ISSN = {2352-7110},
  url = {http://dx.doi.org/10.1016/j.softx.2026.102865},
  DOI = {10.1016/j.softx.2026.102865},
  journal = {SoftwareX},
  publisher = {Elsevier BV},
  author = {Scala,  Francesco and Mandarino,  Francesco and Martirano,  Liliana and Pontieri,  Luigi},
  year = {2026},
  pages = {102865}
}

@article{Skoric2025,
  title = {{Generative Reconstruction of Multimodal Cardiac Waveforms From a Single Vibrational Cardiography Sensor}},
  volume = {29},
  ISSN = {2168-2208},
  url = {http://dx.doi.org/10.1109/JBHI.2025.3561071},
  DOI = {10.1109/jbhi.2025.3561071},
  number = {9},
  journal = {IEEE Journal of Biomedical and Health Informatics},
  publisher = {Institute of Electrical and Electronics Engineers (IEEE)},
  author = {Skoric,  James and D’Mello,  Yannick and Plant,  David V.},
  year = {2025},
  month = sep,
  pages = {6576–6587}
}

@article{oord2016wavenet,
  title={{WaveNet: A Generative Model for Raw Audio}},
  author={Oord, Aaron van den and Dieleman, Sander and Zen, Heiga and Simonyan, Karen and Vinyals, Oriol and Graves, Alex and Kalchbrenner, Nal and Senior, Andrew and Kavukcuoglu, Koray},
  journal={arXiv preprint arXiv:1609.03499},
  year={2016}
}

@inproceedings{stoller2018wave,
  title={{Wave-U-Net: A Multi-Scale Neural Network for End-to-End Audio Source Separation}},
  author={Stoller, Daniel and Ewert, Sebastian and Dixon, Simon},
  booktitle={19th International Society for Music Information Retrieval Conference (ISMIR)},
  year={2018}
}

@inproceedings{lahdenoja2016heart,
  title={Heart rate variability estimation with joint accelerometer and gyroscope sensing},
  author={Lahdenoja, Olli and Humanen, Tero and Tadi, Mojtaba Jafari and P{\"a}nk{\"a}{\"a}l{\"a}, Mikko and Koivisto, Tero},
  booktitle={2016 Computing in Cardiology Conference (CinC)},
  pages={717--720},
  year={2016},
  organization={IEEE}
}

@article{Fye1994,
  title = {A History of the origin,  evolution,  and impact of electrocardiography},
  volume = {73},
  ISSN = {0002-9149},
  url = {http://dx.doi.org/10.1016/0002-9149(94)90135-X},
  DOI = {10.1016/0002-9149(94)90135-x},
  number = {13},
  journal = {The American Journal of Cardiology},
  publisher = {Elsevier BV},
  author = {Fye,  W. Bruce},
  year = {1994},
  month = May,
  pages = {937–949}
}

@misc{who2025cvd,
  author       = {{World Health Organization}},
  title        = {Cardiovascular diseases ({CVDs})},
  howpublished = {WHO Fact Sheet},
  month        = jul,
  year         = {2025},
  note         = {Available: \url{https://www.who.int/news-room/fact-sheets/detail/cardiovascular-diseases-(cvds)}, accessed Sep. 14, 2026}
}

@article{Dai2025,
  title = {{Deep Learning Model Compression With Rank Reduction in Tensor Decomposition}},
  volume = {36},
  ISSN = {2162-2388},
  url = {http://dx.doi.org/10.1109/TNNLS.2023.3330542},
  DOI = {10.1109/tnnls.2023.3330542},
  number = {1},
  journal = {IEEE Transactions on Neural Networks and Learning Systems},
  publisher = {Institute of Electrical and Electronics Engineers (IEEE)},
  author = {Dai,  Wei and Fan,  Jicong and Miao,  Yiming and Hwang,  Kai},
  year = {2025},
  month = Jan,
  pages = {1315–1328}
}

@article{cao2017tensor,
  title={{Tensor Regression Networks with various Low-Rank Tensor Approximations}},
  author={Cao, Xingwei and Rabusseau, Guillaume},
  journal={arXiv preprint arXiv:1712.09520},
  year={2017}
}

\end{document}